\documentclass[sigconf]{acmart}

\usepackage{algorithm,algpseudocode}
\usepackage{multirow}
\usepackage{tabularx}

\newdimen{\algindent}
\newcommand{\commentsymbol}{//}
\algrenewcommand\algorithmiccomment[1]{\hfill \commentsymbol{} #1}
\newdimen{\algindent}
\usepackage{varwidth}

\setcopyright{acmcopyright}
\copyrightyear{2022}
\acmYear{2022}
\setcopyright{rightsretained}
\acmConference[MOBILESoft '22]{IEEE/ACM 9th International Conference on Mobile Software Engineering and Systems}{May 17--24, 2022}{Pittsburgh, PA, USA}
\acmBooktitle{IEEE/ACM 9th International Conference on Mobile Software Engineering and Systems (MOBILESoft '22), May 17--24, 2022, Pittsburgh, PA, USA}
\acmDOI{10.1145/3524613.3527807}
\acmISBN{978-1-4503-9301-0/22/05}

\newcommand{\QD}{QuickDraw}
\newcommand{\dataName}{DoodleUINet}
\newcommand{\toolName}{PSDoodle}

\begin{document}

\title{\toolName{}: Searching for App Screens via Interactive Sketching}

\author{Soumik Mohian}
\affiliation{%
  \department{Computer Science and Engineering Department}
  \institution{University of Texas at Arlington}
  \streetaddress{Box 19015}
  \city{Arlington}
  \state{Texas}
  \country{USA}
  \postcode{76019}
}
\email{soumik.mohian@mavs.uta.edu}

\author{Christoph Csallner}

\affiliation{%
  \department{Computer Science and Engineering Department}
  \institution{University of Texas at Arlington}
  \streetaddress{Box 19015}
  \city{Arlington}
  \state{Texas}
  \country{USA}
  \postcode{76019}
}
\email{csallner@uta.edu}

\begin{abstract}

Keyword-based mobile screen search does not account for screen content and fails to operate as a universal tool for all levels of users. Visual searching (e.g., image, sketch) is structured and easy to adopt. Current visual search approaches count on a complete screen and are therefore slow and tedious.  \toolName{} employs a deep neural network to recognize partial screen element drawings instantly on a digital drawing interface and shows results in real-time. \toolName{} is the first tool that utilizes partial sketches and searches for screens in an interactive iterative way. \toolName{} supports different drawing styles and retrieves search results that are relevant to the user's sketch query.  A short video demonstration is available online at: \textbf{\url{https://youtu.be/3cVLHFm5pY4}}

\end{abstract}




\maketitle

\section{Introduction}

App screen examples help software engineers to accumulate requirements, understand current trends, and motivate to develop a compelling mobile app~\cite{herring2009designpractise,eckert2000sourceinspiration}.  An effective mobile search tool might have a positive impact to keep up with the extensive and increasing use of mobile apps in day-to-day life~\cite{ines2017evalmobileinterface}.

Professional designers search for example screens via keywords through various websites including Google, Dribbble\footnote{\url{https://dribbble.com/}, accessed January 2022.}, and Behance\footnote{\url{https://www.behance.net/}, accessed January 2022.}~\cite{bunian2021vins}. Keyword-based search tools consider stylistic features (e.g., color, style) or image metadata (e.g, date, location, shapes) but fail to consider the contents of the screen~\cite{lee2010designing}. Moreover, novice users often fail to formulate good keywords and thus fail to get the desired search results~\cite{herring2009designpractise}. 

Visual (e.g., image or sketch) search methods are easy, quick to adopt~\cite{yeh2009sikuli}, and commonly used during early software development phases~\cite{Landay95Interactive, Newman99Sitemaps, Wong92Rough, Campos2007Practitioner}. Dependency on a complete query screen makes the iterative nature of the development process tedious (e.g., as in SWIRE~\cite{huang2019swire} or VINS~\cite{bunian2021vins}). Long preprocessing pipelines for a pen on paper approach such as SWIRE include taking a snap, denoising, and projection correction.

\toolName{} is designed for a novice user who cannot or does not want to develop a complete screen at an early phase of software development~\cite{psdoodleMainPaper}. \toolName{} allows a user to draw one UI element at a time.  \toolName{} provides an interactive drawing interface with support for mouse and touch screen devices. For user interaction, the drawing interface includes basic features such as redo, undo stroke, and remove the last icon. 

\begin{figure*}
  \includegraphics[width=\textwidth]{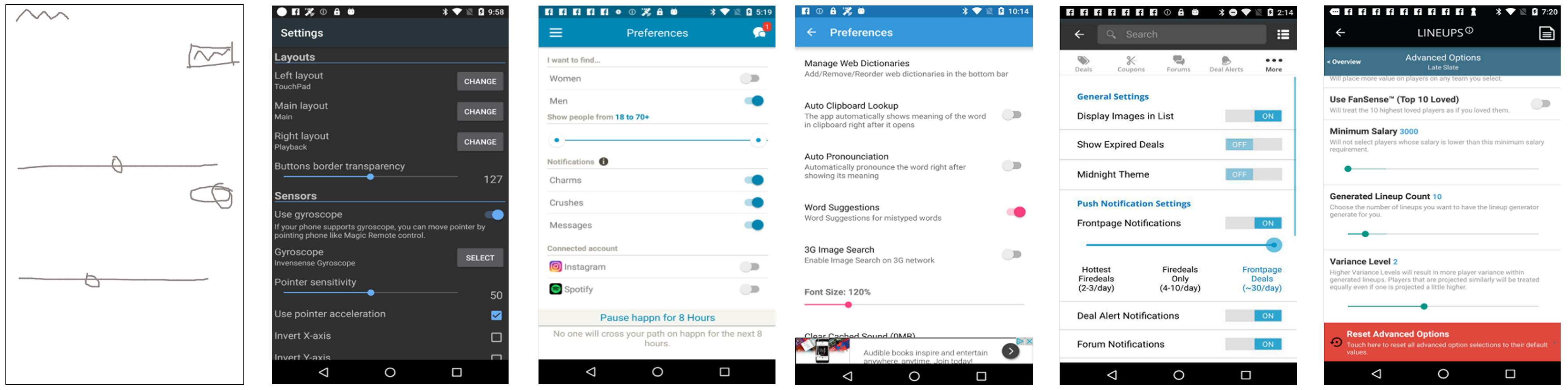}
  \caption{Example mobile app screen search user query (left) and \toolName{}'s top five search results out of 58k app screenshots. The result screens contain all sketched UI elements at about the location they appear in the query sketch.}
  \label{fig:teaser}
\end{figure*}

With each stroke on the drawing interface, \toolName{} utilizes a deep neural network to classify the current UI element and instantly provides a top-3 classification with classification confidence scores. When a user finishes drawing the current UI element by pressing the ``icon done''  button or ``d'' on the keyboard, \toolName{} searches through 58k Rico~\cite{deka2017rico} screens to fetch UI examples based on UI element type, position, and element shape as shown in Figure~\ref{fig:teaser}. \toolName{} fetches 80 screens and displays them at the bottom of the page within 2 seconds.

For evaluation, we enlisted ten software developers who had never used \toolName{} before and have no prior UI/UX design training. Participants used \toolName{} to draw a Rico screen until it appears in the top search results. \toolName's top-10 screen retrieval accuracy was comparable to the state-of-the-art full-screen drawing techniques but reduced the drawing time to one half~\cite{psdoodleMainPaper}. 

To summarize, \toolName{} is the first tool that provides an interactive iterative search-by-sketch screen experience.While the technical-track paper describes \toolName{}'s algorithm and evaluation~\cite{psdoodleMainPaper}, this paper adds details about \toolName{}'s implementation, deployment, support for different sketching styles, ability to surface several relevant search result screens, and user survey results.
All \toolName{} source code, processing scripts, training data, and experimental results are \textbf{open-source} under permissive licenses~\cite{soumikmohianuta_psdoodle_repo, soumik_mohian_DoodleUINet}. 
The tool can be freely used at: \textbf{\url{http://pixeltoapp.com/PSDoodle/}}

\section{Background}

\toolName{} operates on Rico mobile app screens collected from 9.3k Android apps~\cite{deka2017rico}. For each UI element in each screenshot, Rico includes the Android class name, textual information, x/y coordinates, and visibility. Liu et al. collected 73k Rico screen elements and classified all screen components in Rico into 25 UI component kinds (e.g., checkbox, icon, image, text, text button), 197 text button ideas (e.g., no, login, ok), and 135 icon classes (e.g., add, menu, share, star)~\cite{liu2018learning}. \toolName{} supports several frequent Android UI elements reported by Liu et al. \toolName{} utilizes all the hierarchy information and clustering information to find similarity with a user drawing and shows Rico screens as a search result.

\begin{figure}[h!t]
 \centering
 \includegraphics[width=\linewidth,trim={.2in .3in .2in .3in},clip] {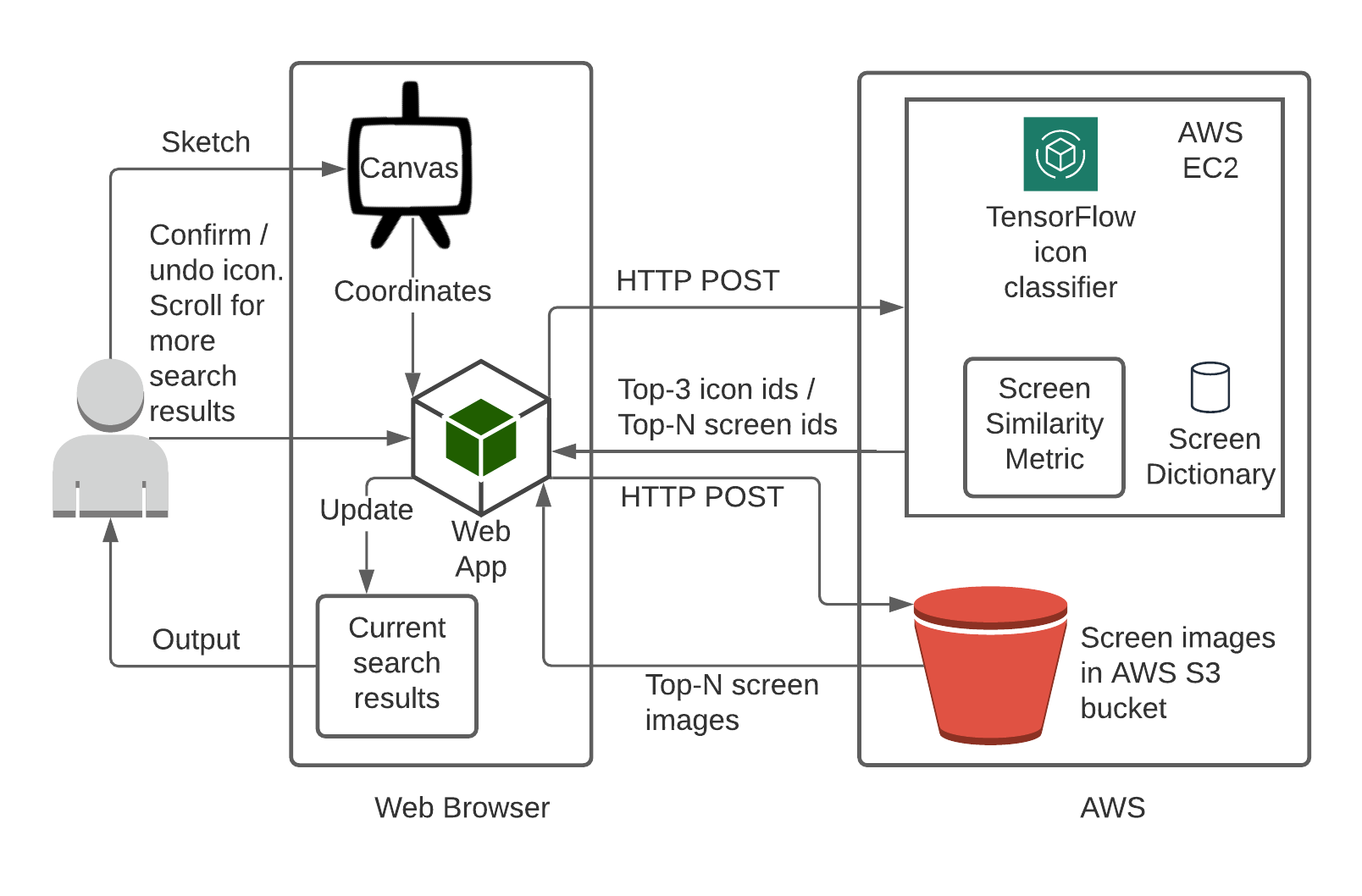}
 \caption{\toolName{} architecture: The user sketches on the \toolName{} webpage (from \url{http://pixeltoapp.com/PSDoodle}). The webpage communicates with the \toolName{} back-end hosted in AWS.}

 \label{fig:tool_flow}
\end{figure}

SWIRE~\cite{huang2019swire} is the most closely related to our work. SWIRE consists of 3.8k scanned low-fidelity pen on paper drawings, each mimicking a complete Rico app screen. All drawings consist of pre-defined conventions with placeholders for image and text (e.g., square borders plus diagonals for an image). We adopted the placeholder techniques in \toolName{} for image, text, and non-supported icons.  SWIRE's deep neural network achieves a top-10 screen retrieval accuracy of 61\%. SWIRE's follow-up work reported a top-10 accuracy of 90.1\%~\cite{sain2020cross}. To generate search results, a user has to draw within a marker, take a snap of the sketch, and provide it to a long pipeline (e.g., de-noising, camera angle correction, and projection correction). For any change in the sketch, a user will most likely have to create a new complete screen sketch from scratch, scan it, and feed it to the processing stages.

\toolName{} incorporates datasets from Google's Quick, Draw~\cite{jongejan2016quick} (``\QD{}'') and \dataName{}~\cite{soumik_mohian_DoodleUINet} to train a deep neural network for recognizing a sketched UI element. \QD{} offers some 50M doodles of 345 everyday categories, from ``aircraft carrier'' to ``zigzag''. \dataName{} gives 11k  crowdworker-created doodles of 16 common Android UI element categories.  \QD{} and \dataName{} represent each doodle as a stroke sequence. Each stroke consists of a series of straight lines drawn from the start to the end of a touch event (e.g., mouse button press and un-press) and each straight line is represented by the x/y coordinates of the endpoints. As an example of how such UI element doodles look like, the left screen in Figure~\ref{fig:teaser} contains six~UI element doodles---two sliders, a switch, a square, and two squiggles.

\section{Overview and Design}

Figure~\ref{fig:tool_flow} shows \toolName{}'s overall architecture. \toolName{}'s website provides a canvas for drawing. The canvas collects the stroke information (x/y coordinates) from a user drawing and the website sends this information to Amazon AWS. \toolName{} uses a deep neural network trained using \QD{}’s network architecture~\cite{quickdraw_rnn,psdoodleMainPaper}. From the stroke information, a Tensorflow implementation of \toolName{}'s neural network hosted on an AWS EC2 node predicts the current drawing's icon category.

The web browser sends all the information rendered on canvas to the Amazon EC2 instance after a user completes drawing a UI element by pressing the ''icon done'' button.  \toolName{}'s similarity metric uses element category, shape, position, and occurrence frequency to seek up the top-N matching screens in its dictionary of Rico screen hierarchies~\cite{psdoodleMainPaper}. After performing the similarity calculation, \toolName{} sends the retrieved screen names to the website. The website fetches the necessary screens from an AWS S3 bucket and displays them to the user.

\subsection{\toolName{}'s Website}

\begin{figure}[h!t]
 \centering
 \includegraphics[width=.8\linewidth,trim={0 0.1cm 0 0.1cm},clip] {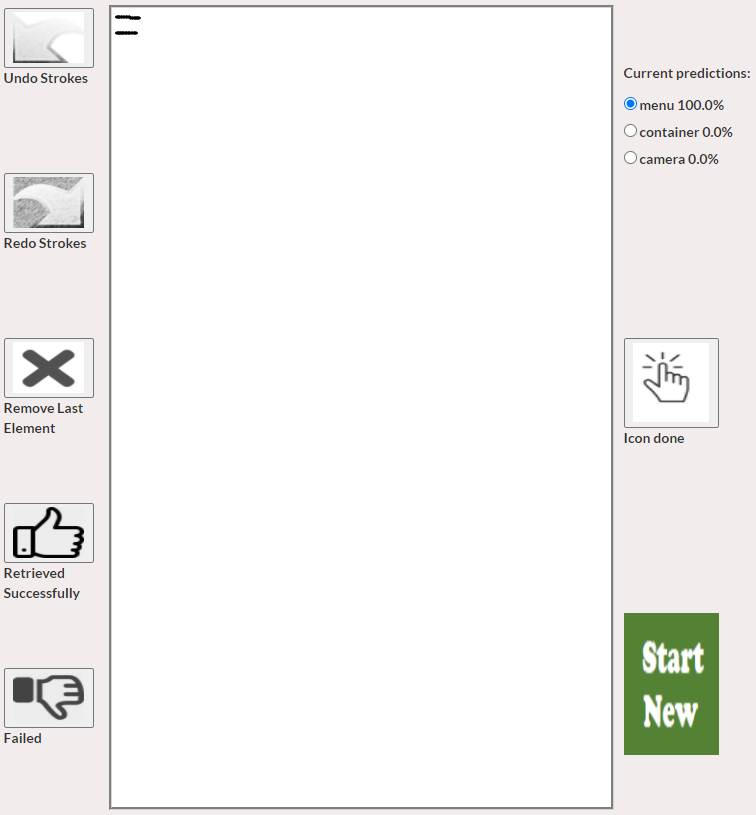}
 \caption{\toolName{} drawing UI, under which \toolName{} shows its current top-N Android search result screens (omitted).}
 \label{fig:searchInterface}
\end{figure}

\toolName{}'s website is hosted on an AWS EC2 general purpose instance (t2.large) with two virtual CPUs and 8 GB of RAM.  \toolName{} provides a guided interactive tutorial (\url{http://pixeltoapp.com/toolIns/})  for first-time users and a cheat sheet of supported UI elements in the top-left of the main page.  To make these instructions self-explanatory, we recruited two UI designers via the freelancing website Upwork\footnote{\url{https://www.upwork.com/}, accessed January 2022.}, recorded their tool usage and incorporated their feedback. This feedback has yielded, e.g., uniform text, position, and size of \toolName{}'s buttons (Figure~\ref{fig:searchInterface}) plus a refactored cheat sheet (e.g., on when and how to draw a text button, Figure~\ref{fig:all_element_categories}).

\toolName{}'s drawing interface is a canvas element with support for both mouse and touch. \toolName{} provides basic drawing features for interactions. Users can undo or redo strokes and remove the last icon (Figure~\ref{fig:searchInterface} top left). \toolName{}'s client-side JavaScript handles all the basic events like touch-start, touch-end, and a stack of strokes to handle redo/undo features.  Each time the user adds a stroke to the current icon doodle, the \toolName{} website detects the touch-end event and sends an HTTP post request with the stroke coordinates to the AWS EC2. \toolName{}'s web application parses the response from AWS EC2 and shows its current top-3 three icon predictions (top right). A user can pick any of these three (and tap ``Icon done'' or 'd' on the keyboard) or continue editing the current icon doodle.

After the user adds (or removes) an icon, the website submits a search query containing all recognized UI elements currently on the canvas plus their on-canvas locations.  Based on \toolName{}'s similarity metric~\cite{psdoodleMainPaper} the website retrieves the top-80 screens' ids from AWS EC2. The website then retrieves the screens corresponding to these ids from the AWS S3 bucket for display on the bottom of the canvas. When a user clicks on a search result, it shows an enlarged version of the screen on the website. A user can navigate between the next 80 search results via the ``next''/``previous'' buttons. The user can give feedback to \toolName{} via the thumbs up/down buttons (lower left) and start a new screen search (lower right).

\toolName{} stores two resolutions of each Rico screen in an AWS S3 bucket. \toolName{} uses the low-resolution images to quickly show the search results (in a gallery view). When a user clicks on an image in the search gallery, \toolName{} fetches the higher-resolution image from the S3 Bucket and displays it of the left side on the website.

\subsection{Recognizing Individual UI Elements}

\begin{figure}[h!t]
 \centering
 \includegraphics[width=\linewidth,trim={0 0.5cm 0 0},clip] {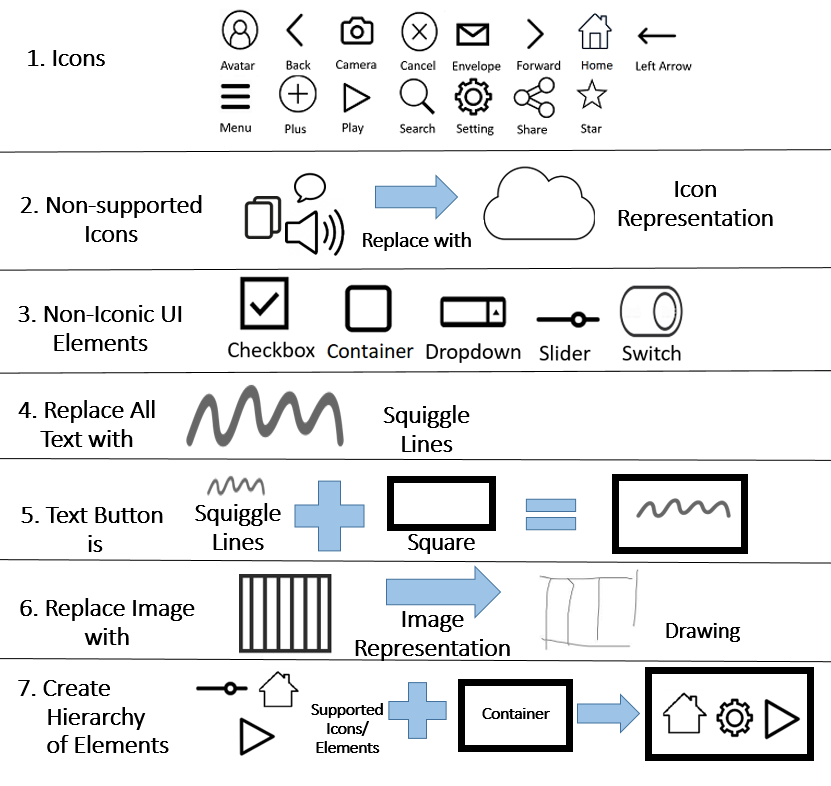}
 \caption{Cheat sheet \toolName{} shows to users: 23 graphical primitives plus compound and nested UI elements.}
 \label{fig:all_element_categories}
\end{figure}

Figure~\ref{fig:all_element_categories} gives an overview of \toolName{}'s visual query language, i.e., the 23 graphical primitives \toolName{} recognizes and how the user can combine them to create compound and nested UI elements. According to the number of element labels inferred by Liu et al.~\cite{liu2018learning}, \dataName{}~\cite{soumik_mohian_DoodleUINet} covers several of the most popular UI elements in Rico. \toolName{} uses 7 \QD{} classes that were a suitable match for UI sketching. \toolName{} provides placeholders for text (squiggle line), image (jailwindow), and for not directly supported icons (cloud). Placeholders help the user to avoid fine details of text and images. 

\toolName{} provides options to draw compound UI elements. For example, the Android text button is one ``text'' inside a ``square'' drawn separately. Sketching one UI element at a time permits users to nest UI elements within a container by drawing them separately. 

\toolName{} uses a deep neural network similar to \QD{}'s network architecture~\cite{quickdraw_rnn} to identify UI element class from strokes. Three 1-D convolutional neural networks (CNN) layers followed by three Bi-LSTM layers, and a fully-connected layer make up \toolName{}'s deep neural network. \toolName{} used \dataName{} (600 doodles for each of its 16 classes) and a random 600-doodle sample of each of \toolName{}'s 7 \QD{} classes to train the deep neural network and yielded a 94.5\% test accuracy.

We adopted a faster TensorFlow implementation of the deep neural network in AWS EC2 that avoids regenerating the network graph for predicting class labels for each stroke.   \toolName{} detects the class label, confidence score with the Tensorflow implementation, and shows them instantly (in under a second). \toolName{} guides a user to express their drawing intention by instantly showing the updated prediction with each stroke.

\section{Exploring \toolName{}'s Usage}

Following the most closely related work~\cite{huang2019swire,sain2020cross}, we evaluated screen search performance by measuring top-k (screen) retrieval accuracy~\cite{psdoodleMainPaper}. We thus showed a participant a target screen to sketch and measured where in the result ranking the target screen appears. Top-k retrieval accuracy is the most common metric for sketch-based image retrieval tasks and correlates with user satisfaction~\cite{huffman2007well}.

While \toolName{}'s top-10 screen retrieval accuracy of 88\% is similar to the state-of-the-art's 90\%, \toolName{} cuts the state-of-the-art's screen retrieval time in half~\cite{psdoodleMainPaper}. In this section, we explore how \toolName{} supports different sketching styles, how many of the tool's top-10 search results are relevant to the user's query sketch, and how users have described \toolName{}.

\subsection{Supporting Different Sketching Styles}

Table~\ref{tab:confindenceStroke} shows UI category drawings with different numbers of strokes in the test dataset, showing a variety of \QD{}'s game participants' and \dataName{} crowdworkers' drawing styles. \toolName{} can thus detect sketches with high confidence for different drawing styles. 
 
As an example, in all cases when \toolName{} classified a slider test sketch that consists of five strokes, it assigned a 100\% confidence that the sketch is indeed a slider. But \toolName{} similarly had correctly classified (with high confidence) other slider test sketches, e.g., those consisting of four or three strokes. The likely reason for this behavior is that \toolName{}'s classifier has been trained on sketches from a variety of people. Similarly, adding more training samples from a wider variety of crowdworkers may make \toolName{} even more robust to different drawing styles.

\begin{table}[htbp]
\begin{center}
  \caption{Average confidence \toolName{} has in a sketch of the given total stroke count belonging to sketch's target category. For example, the network has 50\% confidence on average for completed avatar sketches of a total of 7~strokes to be avatars.}

  \label{tab:confindenceStroke}
  \begin{tabular}{l|rrrrrrrrr}
   \hline
    \multicolumn{1}{c|} {\textbf{Cat.}} &
    \multicolumn{9}{c} {\textbf{Confidence by sketch's total strokes}} \\
     & 1 & 2 & 3 & 4 & 5 & 6 & 7 & 8 & 9+  \\
       \hline

Camera & - & 92 & 96 & 96 & 100 & 100 & - & 100 & -\\
Cloud & - & - & 75 & 97 & 95 & 100 & 89 & 89 & 100 \\ 
Envel. & 100 & 96 & 100 & 100 & 86 & 100 & - & - & 100 \\ 
House & 80 & 97 & 94 & 100 & 72 & 100 & 100 & 100 & 50 \\
Jail-win & 89 & 78 & 100 & 100 & 67 & 0 & - & - & -\\
Square & 98 & 100 & 100 & 75 & - & - & - & - & -\\
Star & 99 & 100 & 100 & 100 & 100 & - & 100 & - & -\\       
       
       \hline       
Avatar & - & 100 & 82 & 96 & 89 & 100 & 50 & 100 & - \\
Back & 97 & 100 & 0 & - & - & 100 & - & - & -  \\
Cancel & - & 100 & 77 & 50 & 50 & - & - & - & - \\
Checkb. & 100 & 96 & 61 & 71 & 83 & 0 & 100 & 100 & 67 \\
Drop-d. & - & 100 & 98 & 88 & 100 & 100 & 100 & 100 & 100 \\
Forward & 100 & 100 & - & - & - & - & - & - & - \\ 
Left-arr. & 80 & 87 & 92 & - & 100 & - & 50 & - & - \\
Menu & - & 100 & 100 & 100 & 100 & 100 & - & - & - \\
Play & 96 & 96 & 95 & 100 & 100 & - & - & - & 0 \\ 
Plus & - & 100 & 93 & 100 & 100 & - & 100 & 0 & - \\ 
Search & 100 & 98 & 100 & 100 & 100 & - & - & - & 100 \\ 
Setting & 100 & 94 & 92 & 67 & 100 & 50 & - & 100 & 83 \\ 
Share & - & 0 & - & - & 100 & 100 & 98 & 100 & 100 \\ 
Slider & 100 & 94 & 97 & 100 & 100 & 100 & 100 & - & - \\ 
Squiggle & 98 & 83 & 100 & 100 & 100 & 100 & - & - & 0 \\ 
Switch & 100 & 97 & 80 & 91 & 100 & 100 & 100 & 50 & 0 \\ 
 \hline
\end{tabular}
\end{center}
\end{table}

\subsection{Surfacing Several Relevant Result Screens}

Figure~\ref{fig:teaser} shows an example partial screen sketch and \toolName{}'s top 5 search results. The search results are of high quality as the result screens contain all sketched UI elements at about the location they appear in the query sketch. 

In a user study with 10 participants\footnote{We recruited 10 Computer Science students who had no prior UI/UX design experience. Each participant first spent on average some 9 minutes in \toolName{}'s interactive tutorial (\url{http://pixeltoapp.com/toolIns/}). We then gave Rico target screens to sketch.} we received similar feedback. For each of a total of 34 screen sketches, participants judged the quality of each of \toolName{}'s top-10 result screens. Participants judged 145/340 of these screens as relevant to their search query.

\subsection{User Experience}

Nine users have filled out a brief survey about their experience, including the following two open-ended questions.

\begin{description}
  \item[Q1] What improvement/features do you suggest to make the interface better for a user?
  \item[Q2] How was the overall search results?
\end{description}

While all answers are available\footnote{\url{https://github.com/soumikmohianuta/PSDoodle/blob/master/ComparisonResult/ToolSurvey/PSDoodle_Tool_Survey.pdf}}, following are a few highlights. Regarding improvements (Q1), users asked for more icon support. Supporting more UI element categories is mostly a matter of gathering additional training samples and retraining the deep neural network. Regarding the overall search results (Q2), users were generally positive. Following is one quote:

``good, the results were similar to what I was looking for''.

Another user said the following.
\begin{quote}
``I tested different shapes the overall result was good. [..] I was overall satisfied with what I tested.''
\end{quote}

\section{Conclusions}

Current approaches to searching through existing repositories are either slow or fail to address the need of novice users. Interactive partial sketching, which is more structured than a keyword search and faster than complete-screen inquiries, is a viable option. \toolName{} is the first tool to offer interactive screen search with sketching and live search results.

\begin{acks}
Christoph Csallner has a potential research conflict of interest due to a financial interest with Microsoft and The Trade Desk. A management plan has been created to preserve objectivity in research in accordance with UTA policy. This material is based upon work supported by the National Science Foundation (NSF) under Grant No. 1911017.
\end{acks}

\bibliographystyle{ACM-Reference-Format}
\bibliography{main}

\appendix

\end{document}